\documentclass[11pt]{article}
\PassOptionsToPackage{table}{xcolor}
\usepackage[preprint]{acl}
\usepackage[table]{xcolor}
\usepackage{algorithm}
\usepackage{algpseudocode}
\usepackage{times}
\usepackage{latexsym}
\usepackage[T1]{fontenc}
\usepackage[utf8]{inputenc}
\usepackage{microtype}
\usepackage{inconsolata}
\usepackage{graphicx}
\usepackage{booktabs}
\usepackage{amsmath,amssymb}
\usepackage{amsthm}
\theoremstyle{definition}
\newtheorem{definition}{Definition}
\newtheorem{assumption}{Assumption}
\newtheorem{problem}{Problem}
\theoremstyle{plain}
\newtheorem{proposition}{Proposition}
\newcommand{\Ready}{\textnormal{\textsc{Ready}}}
\newcommand{\Blocked}{\textnormal{\textsc{Blocked}}}
\newcommand{\Other}{\textnormal{\textsc{Other}}}

\title{RadixFlow: Filling Execution Bubbles in Prefix-Cached Hybrid LLM Serving}
\author {
  \begin{minipage}{\textwidth}
  \centering
    {\bf Huihao Jing}\textsuperscript{ \hspace{-0.2em}\includegraphics[height=1em]{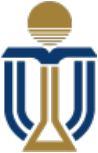}},
    {\bf Wenbin Hu}\textsuperscript{ \hspace{-0.2em}\includegraphics[height=1em]{figs/HKUST.pdf}},
    {\bf Shaojin Chen}\textsuperscript{ \hspace{-0.2em}\includegraphics[height=1em]{figs/HKUST.pdf}},
    {\bf Haochen Shi}\textsuperscript{ \hspace{-0.2em}\includegraphics[height=1em]{figs/HKUST.pdf}},
    {\bf Haoyu Huang}\textsuperscript{ \hspace{-0.2em}\includegraphics[height=1em]{figs/HKUST.pdf}},\\
    {\bf Sirui Zhang}\textsuperscript{ \hspace{-0.2em}\includegraphics[height=1em]{figs/HKUST.pdf}},
    {\bf Yuxuan Liu}\textsuperscript{ \hspace{-0.2em}\includegraphics[height=1em]{figs/HKUST.pdf}},
    {\bf Zhongwei Xie}\textsuperscript{ \hspace{-0.2em}\includegraphics[height=1em]{figs/HKUST.pdf}},
    {\bf Haoran Li}\textsuperscript{
      \hspace{-0.2em}
      \includegraphics[height=1em]{figs/HKUST.pdf}
    }\footnotemark[2]\hspace{0.15em}\\
    \textsuperscript{\includegraphics[height=1em]{figs/HKUST.pdf}}HKUST \\
    \texttt{hjingaa@connect.ust.hk}\\
  \end{minipage}
}
\begin{document}
\maketitle
\renewcommand{\thefootnote}{\fnsymbol{footnote}}
\footnotetext[2]{Corresponding author.}
\renewcommand{\thefootnote}{\arabic{footnote}}
\setcounter{footnote}{0}
\begin{abstract}
Prefix reuse improves LLM inference by sharing the KV cache of common prompt prefixes. This helps only if the scheduler can execute a matched branch in the current worker round. Schedulers assume that a matched prefix is ready, which holds for full attention but fails for hybrid LLMs. There, a branch (the consumer) may need recurrent state that another branch (the producer) is still computing in the same round. The native scheduler stops its waiting-queue scan at this blocked consumer, and the resulting head-of-line blocking creates \emph{execution bubbles}: idle batch slots that independent work could fill. We introduce RadixFlow, which decouples prefix matching from state readiness. Its state-ready traversal turns a global scan barrier into a local wait: only the consumer is blocked until the next round, while feasible independent branches keep filling the batch under the native state check. On five hybrid models, RadixFlow lowers mean latency by 3\% to 24\% and worker rounds by 11\% to 44\% at almost unchanged reuse. On the larger Qwen3.5 models, ablations show that most of the gain comes from traversal, which adds little on a full-attention control. Heavier rounds limit how much of this saving reaches latency. 
\par All techniques have been implemented and integrated into
\href{https://github.com/mode-io/vllm-jev}{vllm-jev}.

\end{abstract}

\section{Introduction}
Efficient LLM inference in online settings increasingly involves long shared contexts. Retrieval-augmented generation, multi-turn dialogue, and agent pipelines send queries that repeat the same documents, system prompts, or histories, placing inference in a prefill-dominated regime \citep{dexter2025scheduling,patel2023splitwise}. An important approach to improving efficiency in this regime is prefix reuse, which keeps the KV cache of shared prefixes \citep{kwon2023pagedattention,zheng2024sglang,gim2024promptcache,yu2023pensieve}. Under a stream of queries, however, its benefit depends on how the scheduler groups the remaining work into worker rounds \citep{zhong2024distserve}.

\begin{figure*}[!t]
\centering\includegraphics[width=\textwidth]{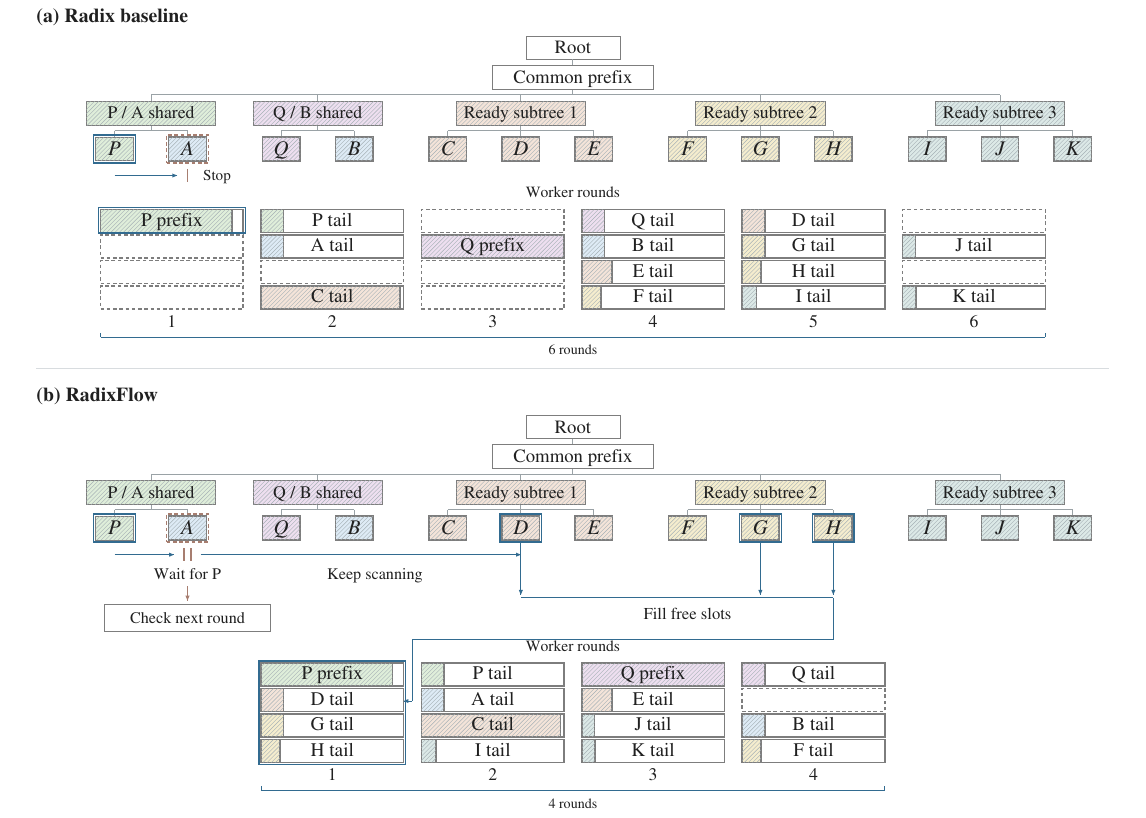}
\caption{Consumer $A$ waits for $P$, while independent $D$ can still enter the round. Both panels share the same prefix structure and fifteen work fragments. The baseline stops at consumer $A$, whose state depends on producer $P$; RadixFlow blocks only $A$ and keeps scanning. Blue outlines mark first-round selections, dashed cells are empty, and shaded lengths show token load. Blocked consumers are woken for a recheck in the next round. These idealized schedules process the same 12,970 tokens in six versus four rounds.}
\label{fig:mechanism}
\end{figure*}

Existing schedulers take one step for granted: a branch that matches a cached prefix can run. This holds for full attention, where a matched prefix needs only its KV pages \citep{kwon2023pagedattention,zheng2024sglang}. Hybrid LLMs break it, because resuming a branch also requires recurrent and convolutional state at the match boundary \citep{gu2023mamba,dao2024ssm,yang2024gdn,pan2025marconi,sglang2026unified}. Following operating-system terminology, we call the branch that computes this state the \emph{producer} and a branch that must read it the \emph{consumer}. When both fall in the same worker round, the consumer has a \emph{same-round dependency}: its prefix is known, but its state is readable only after the producer executes. In the evaluated vLLM path, the consumer's allocation failure ends the waiting-queue scan \citep{vllm2026cache,vllm2026scheduler}, causing \emph{head-of-line blocking}. In Figure~\ref{fig:mechanism}, $A$ must wait for $P$, so the scan also excludes the independent branch $D$, although $D$ passes the native state check and fits the remaining budget. We call such legally fillable idle capacity an \emph{avoidable execution bubble}.

We introduce RadixFlow, a batch-construction policy that separates a prefix match from state readiness. Its state-ready traversal blocks only the consumer, which waits in a wait queue and is woken for a recheck in the next round, while independent branches join the current batch. Other allocation outcomes keep their native handling. Flexible admission control briefly collects arrived work before the scan. In any fixed round, RadixFlow selects a superset of the native batch. On five hybrid models, RadixFlow lowers mean client latency by 3\% to 24\% while prefix reuse stays within 0.03 percentage points of Radix. Traversal contributes most of the gain on the larger Qwen3.5 models and little on a full-attention control. The remaining rounds carry more work, so their cost decides how much of the 11\% to 44\% round reduction reaches latency \citep{agrawal2024sarathi,pan2025fasttree}.

Our contributions are:
\begin{itemize}
\item \textbf{A cache hit is not a runnable branch.} We identify same-round dependencies in hybrid LLM serving, formalize readiness and avoidable execution bubbles, and show how head-of-line blocking leaves feasible work unscheduled despite a prefix match and spare capacity.
\item \textbf{Block the consumer, not the queue.} We develop state-ready traversal, which blocks only the consumer and keeps selecting independent work, and prove its fixed-round feasibility and selection properties without changing cache storage or native state checks.
\item \textbf{Fewer rounds, faster serving.} RadixFlow lowers mean latency by 3\% to 24\% on five hybrid models at nearly unchanged reuse. Ablations, round-occupancy traces, and worker-cost accounting relate the gains to batch construction.
\end{itemize}

\section{Related Work}
Prefix caching decides what state can be reused. PagedAttention shares KV pages, RadixAttention indexes exact prefixes, and hybrid caches retain the recurrent components needed at a reuse boundary \citep{kwon2023pagedattention,zheng2024sglang,pan2025marconi,sglang2026unified}. RadixFlow preserves that storage policy and uses the native state check to select executable work.

Scheduling systems balance reuse, load, and waiting through request ordering or batch formation \citep{dexter2025scheduling,srivatsa2025preble,agrawal2024sarathi,zheng2026batchllm}. Bounded dynamic batching motivates our timer controls \citep{triton2025batching}. RadixFlow acts inside a round: a same-round dependency changes which branches are selected, without changing their native order or bypassing a state read. Kernel and tree-aware batching work also motivates measuring per-round cost \citep{dao2022flashattention,pan2025fasttree}. Appendix~\ref{app:extendedrelated} gives the broader comparison.

\section{Problem Formulation}
\label{sec:problem}
We study batch construction for a prefix-cached worker. A branch must match the intended prefix, have usable state at its resume boundary, and fit the remaining resources. These requirements determine which arrived branches can execute together. We follow the finite-stream setting of \citet{dexter2025scheduling}.

\subsection{How a branch executes}
\label{sec:architecture}
\begin{figure*}[t]
\centering\includegraphics[width=\textwidth]{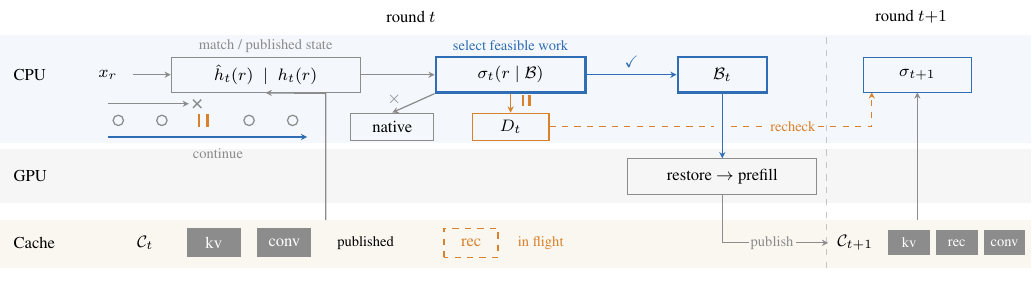}
\vspace{-0.08in}
\caption{RadixFlow changes selection, not native state checks or cache updates. A checkmark selects feasible work for $\mathcal B_t$; a pause blocks a consumer with a same-round dependency in the wait queue $D_t$, and a cross preserves native handling. Solid state components are published. The outlined recurrent component is still in flight and is not part of $\mathcal C_t$; blocked consumers are woken for a recheck in round $t+1$. The queue inset contrasts head-of-line blocking with continuing past the blocked consumer.}
\label{fig:architecture}
\end{figure*}

A branch can reuse state only after publication. Figure~\ref{fig:architecture} links prefix lookup and the native state check to batch construction, restoration, and prefill. Restoration precedes the resumed forward; publication precedes dependent reuse. Independent CPU control can overlap GPU work \citep{nvidia2026timing}, and several branches share one worker invocation \citep{yu2022orca,agrawal2024sarathi}.

\subsection{Streams, state, and rounds}
\paragraph{Notation.}
A worker round holds at most $S$ rows and $T$ scheduled tokens; $x_{\le h}$ denotes the first $h$ tokens of input $x$. Request $q_i$ arrives at offered time $a_i$ and contains input branches $x_{ij}$. Let $\mathcal R$ be all branches; for $r=(i,j)$, write $x_r=x_{ij}$. The policy observes only arrived inputs. In Figure~\ref{fig:mechanism}, $P$ and $A$ are branches sharing a prefix. Full stream notation appears in Appendix~\ref{app:formalmodel}.

\paragraph{State readiness.}
Let $\mathcal C_t$ contain published state that passes the native state check. A hybrid reuse boundary needs KV, recurrent, and convolutional components. Its ready length $h_t(r)$ is the largest usable boundary, while $\hat h_t(r)$ is the longest token match. For $A$, $\hat h_t(A)>h_t(A)$ when $P$ is still producing the matched state. That state enters $\mathcal C_{t+1}$ after execution. Appendix~\ref{app:formalmodel} gives Definition~\ref{def:ready}.

Let $\mathcal A_t$ contain arrived, admitted branches with unscheduled work. A feasible partial batch $\mathcal B\subseteq\mathcal A_t$ fits both budgets and resumes each branch from valid published state or its own earlier state. This permits $D$ to join $P$, but prevents $A$ from reading state that $P$ is still producing.

\begin{definition}[Readiness outcome]
\label{def:outcome}
Fix round $t$ and a partial batch $\mathcal B\subseteq\mathcal A_t$. The native state check maps each $r\in\mathcal A_t\setminus\mathcal B$ to an outcome $\sigma_t(r\mid\mathcal B)\in\{\Ready{},\Blocked{},\Other{}\}$. \Ready{} means $\mathcal B\cup\{r\}$ is feasible. \Blocked{} means the check rejects $r$ because a component at $\hat h_t(r)$ is still being written by some producer $r'\in\mathcal B$, that is, $r$ is a consumer with a same-round dependency. \Other{} covers all remaining outcomes, including budget overflow and unrecognized allocation failures.
\end{definition}
\Ready{} is stronger than a ready boundary: it also requires the branch to fit the round. With $P$ selected, $A$ is \Blocked{}, while an independent branch such as $D$ can be \Ready{}.

\begin{definition}[Execution bubble]
\label{def:bubble}
Unused rows are counted by $b_t=S-|\mathcal B_t|$. An \emph{avoidable execution bubble} exists when $b_t>0$ and some $r\in\mathcal A_t\setminus\mathcal B_t$ has $\sigma_t(r\mid\mathcal B_t)=\Ready{}$: the batch leaves out arrived work that could legally fill its capacity.
\end{definition}
In the first prefix-stop round of Figure~\ref{fig:mechanism}, $P$ occupies one of four rows, leaving three rows empty. This row-based measure differs from pipeline bubbles \citep{agrawal2024sarathi}. Our target is legally fillable capacity; a full token budget or an empty executable set can also leave rows unused.

\begin{problem}[Online batch construction]
\label{prob:batch}
Let $C_i(\pi)$ be the completion time of the last branch of logical request $q_i$. Given arrived work, state $\mathcal C_t$, and the native outcomes $\sigma_t$, construct feasible batches online to minimize mean flow time,
\[
 J(\pi)=\frac{1}{n}\sum_{i=1}^{n}\bigl(C_i(\pi)-a_i\bigr).
\]
\end{problem}
Round count and round cost determine service work, while arrivals and queueing determine when each request receives it. Section~\ref{sec:cost} relates the service quantities; the experiments measure both offered-arrival and post-submission latency.

\subsection{Where bubbles arise}
\label{sec:bubbles}
The evaluated vLLM waiting loop ends its scan when native allocation returns no blocks. Its hybrid-state manager produces this outcome for matched current-step state, even with available pool capacity \citep{vllm2026cache,vllm2026scheduler}. Appendix~\ref{app:baselinepath} identifies the source path. We model this path as scanning $\mathcal A_t$ in native order $\prec$ and adding each \Ready{} branch. We call this rule \emph{prefix-stop} when the scan ends at the first branch whose outcome is not \Ready{}. Consider a producer $P$ and a consumer $A$ that share a prefix. Once $P$ is selected, $\sigma_t(A\mid\mathcal B)=\Blocked{}$. A prefix-stop scan ends at $A$, so every \Ready{} branch $r\succ A$ is left out, which is head-of-line blocking. Continuing the scan pays off when arrived independent work passes the native state check and fits the remaining capacity. In that case, the empty row is avoidable by Definition~\ref{def:bubble}.

Figure~\ref{fig:mechanism} shows the effect. Both schedules contain the same fifteen work fragments and 12,970 tokens. Stopping at $A$ takes six four-slot rounds and leaves nine slots empty. Continuing past $A$ takes four rounds and leaves one slot empty.

\paragraph{Packing reference.}
For fixed work fragments with $W$ tokens, $V$ row visits, and dependency depth $D$, at least $\max\{\lceil W/T\rceil,\lceil V/S\rceil,D\}$ rounds are required. The four-round example attains this capacity bound for its 12,970 tokens and fifteen fragments. Proposition~\ref{prop:bound} and its proof appear in Appendices~\ref{app:formalmodel} and~\ref{app:proofs}.

\section{RadixFlow}
\label{sec:method}
RadixFlow recovers executable capacity within a worker round. Its main rule is state-ready traversal: block only the consumer of a same-round dependency and keep selecting independent work, removing head-of-line blocking. Flexible admission determines when arrived branches become eligible before that scan. Both stages retain the native state check and cache management.

\subsection{State-ready traversal}
\label{sec:batchalgorithm}
Traversal replaces the prefix-stop rule with one change. It scans eligible branches in the native order $\prec$ and keeps a partial batch $\mathcal B$ and a wait queue $D_t$. A branch with $\sigma_t(r\mid\mathcal B)=\Ready{}$ is selected through the native allocator. A branch with outcome \Blocked{} joins $D_t$ and stays registered, and the scan continues. An \Other{} outcome keeps the native action (Algorithm~\ref{alg:round}). The outcome of a branch is read at its scan position, against the batch built so far.

Continuing after every allocation failure would mix different causes. A \Blocked{} consumer waits for an already selected producer; an \Other{} result can indicate a resource limit or another failure. RadixFlow continues in the former case and preserves native handling in the latter. Thus $A$ waits for $P$ while the scan visits $D$.

\begin{algorithm}[t]
\caption{RadixFlow constructs a feasible round by blocking only consumers of same-round dependencies.}
\label{alg:round}
\small
\begin{algorithmic}[1]
\State Apply bounded admission to arrived branches.
\State $\mathcal B\gets\emptyset$; $D_t\gets\emptyset$
\For{eligible branch $r$ in native order $\prec$}
    \State $\sigma\gets\sigma_t(r\mid\mathcal B)$ \Comment{Native state check}
    \If{$\sigma=\Ready{}$}
        \State Select $r$ through the native allocator.
        \State $\mathcal B\gets\mathcal B\cup\{r\}$ \label{alg:accept}
    \ElsIf{$\sigma=\Blocked{}$}
        \State $D_t\gets D_t\cup\{r\}$; keep $r$ registered. \label{alg:defer}
        \State Continue scanning.
    \Else
        \State Apply the native action. \Comment{May stop}
    \EndIf
\EndFor
\State Execute $\mathcal B_t=\mathcal B$; publish to $\mathcal C_{t+1}$.
\State Wake the consumers in $D_t$ for a recheck in round $t+1$.
\end{algorithmic}
\end{algorithm}

\paragraph{Properties.}
For fixed state and native order, RadixFlow constructs a feasible batch containing every branch selected by prefix-stop. If readiness is monotone as the batch grows and no \Other{} outcome stops the scan, no avoidable bubble remains. Proposition~\ref{prop:traversal} in Appendix~\ref{app:formalmodel} states the assumptions and Appendix~\ref{app:proofs} proves the result. This is a fixed-round guarantee; stream-wide savings are measured below.

\paragraph{Cost.}
Continuing the scan makes additional queue visits and repeats the native state check. Each blocked consumer is visited once per pass, so the extra work grows linearly with the number of inspected branches.

\subsection{Flexible admission}
Admission control trades batch size against waiting time. The admission deadline is 5\,ms, about 9\% of the 52.95\,ms native mean worker-call interval on Qwen3.5-0.8B. We fix this short bound across models rather than tune it on evaluation results. A branch is released earlier in three cases: the worker already has running work, the held branches fill the available rows, or their residual token demand fills the token budget. Residual demand is computed from current native cache availability. A branch does not wait for the rest of its request group. Flex denotes flexible admission. The Radix+Flex variant keeps this rule and removes state-ready traversal, so it isolates the effect of admission. The timer controls differ from Radix+Flex because they release work only at their deadline or when rows are full.

\subsection{Fragmentation extension}
A small residual budget can split a branch into an extra fragment. RadixFlow+Frag defers such a branch when the split would shorten its next aligned segment, and then looks for another fit. This can reduce fragmentation, but it can also delay progress. We keep RadixFlow as the main method and test RadixFlow+Frag as an ablation.

\paragraph{Design choices.}
RadixFlow changes selection rather than checkpoint placement. Additional recurrent snapshots can enable more reuse but also incur copying and retention costs; waiting for every branch can postpone work that is already executable. The policy instead uses published native state and releases each arrived branch independently. Appendix~\ref{app:design} details these choices.

\section{Experiments}
\label{sec:setup}
We ask: (Q1) Does RadixFlow improve serving performance? (Q2) Where does the gain come from? (Q3) How does per-round cost shape the latency improvement?

\begin{table*}[t]
\centering\footnotesize
\setlength{\aboverulesep}{0.5pt}\setlength{\belowrulesep}{0.5pt}
\setlength{\tabcolsep}{4pt}\renewcommand{\arraystretch}{1.08}
\begin{minipage}[t]{0.483\textwidth}
\begin{tabular*}{\linewidth}{@{}l@{\extracolsep{\fill}\hspace{5pt}}rrrr@{}}
\toprule
\shortstack[l]{\textbf{Method}\\\rule{0pt}{6pt}} & \shortstack[r]{\textbf{Latency}\\(s) $\downarrow$} & \shortstack[r]{\textbf{Replay}\\(req/s) $\uparrow$} & \shortstack[r]{\textbf{Rounds}\\$K$ $\downarrow$} & \shortstack[r]{\textbf{Reuse}\\(\%) $\uparrow$}\\
\midrule
\multicolumn{5}{@{}l@{}}{\makebox[\linewidth][l]{\textbf{Qwen3.5-0.8B}\hfill{\scriptsize $S{=}16$, $T{=}8192$, 5.0\,req/s}}}\\
\cmidrule(l{0pt}r{0pt}){1-5}
Radix & 0.256 & 4.806 & 19,803 & 81.362 \\
Radix+Timer5 & 0.260 & 4.805 & 19,371 & 81.367 \\
Radix+Timer10 & 0.264 & 4.805 & 19,181 & \textbf{81.373} \\
Radix+Flex & 0.249 & 4.805 & 18,409 & 81.364 \\
\rowcolor{gray!12}
RadixFlow & 0.232 & 4.806 & \textbf{17,610} & 81.366 \\
RadixFlow+Frag & \textbf{0.228} & 4.806 & 17,744 & 81.371 \\
\cmidrule(l{0pt}r{0pt}){1-5}
\addlinespace[1pt]
\multicolumn{5}{@{}l@{}}{\makebox[\linewidth][l]{\textbf{Qwen3.5-4B}\hfill{\scriptsize $S{=}64$, $T{=}8192$, 5.0\,req/s}}}\\
\cmidrule(l{0pt}r{0pt}){1-5}
Radix & 13.1 & 4.210 & 13,424 & 80.405 \\
Radix+Timer5 & 11.8 & 4.269 & 12,698 & 80.341 \\
Radix+Timer10 & 11.5 & 4.274 & 12,526 & 80.377 \\
Radix+Flex & 12.8 & 4.206 & 12,942 & 80.433 \\
\rowcolor{gray!12}
RadixFlow & \textbf{10.0} & 4.387 & \textbf{10,146} & 80.431 \\
RadixFlow+Frag & 10.1 & 4.345 & 10,296 & \textbf{80.450} \\
\cmidrule(l{0pt}r{0pt}){1-5}
\addlinespace[1pt]
\multicolumn{5}{@{}l@{}}{\makebox[\linewidth][l]{\textbf{Qwen3.5-9B}\hfill{\scriptsize $S{=}64$, $T{=}8192$, 5.0\,req/s}}}\\
\cmidrule(l{0pt}r{0pt}){1-5}
Radix & 26.7 & 3.240 & 11,042 & 79.686 \\
Radix+Timer5 & 26.5 & 3.245 & 10,365 & 79.461 \\
Radix+Timer10 & 26.5 & 3.257 & 10,279 & 79.471 \\
Radix+Flex & 26.5 & 3.252 & 10,770 & \textbf{79.757} \\
\rowcolor{gray!12}
RadixFlow & \textbf{24.8} & 3.368 & \textbf{7,444} & 79.696 \\
RadixFlow+Frag & 24.9 & 3.352 & 7,534 & 79.755 \\
\bottomrule
\end{tabular*}
\end{minipage}\hfill
\begin{minipage}[t]{0.483\textwidth}
\begin{tabular*}{\linewidth}{@{}l@{\extracolsep{\fill}\hspace{5pt}}rrrr@{}}
\toprule
\shortstack[l]{\textbf{Method}\\\rule{0pt}{6pt}} & \shortstack[r]{\textbf{Latency}\\(s) $\downarrow$} & \shortstack[r]{\textbf{Replay}\\(req/s) $\uparrow$} & \shortstack[r]{\textbf{Rounds}\\$K$ $\downarrow$} & \shortstack[r]{\textbf{Reuse}\\(\%) $\uparrow$}\\
\midrule
\multicolumn{5}{@{}l@{}}{\makebox[\linewidth][l]{\textbf{Qwen3.5-27B}\hfill{\scriptsize $S{=}64$, $T{=}8192$, 5.0\,req/s}}}\\
\cmidrule(l{0pt}r{0pt}){1-5}
Radix & 164 & 0.743 & 11,748 & 68.497 \\
Radix+Timer5 & 168 & 0.728 & 10,524 & 67.426 \\
Radix+Timer10 & 169 & 0.726 & 10,453 & 67.334 \\
Radix+Flex & 163 & 0.746 & 11,555 & 68.567 \\
\rowcolor{gray!12}
RadixFlow & 159 & 0.766 & 6,529 & 68.508 \\
RadixFlow+Frag & \textbf{151} & 0.805 & \textbf{6,293} & \textbf{70.166} \\
\cmidrule(l{0pt}r{0pt}){1-5}
\addlinespace[1pt]
\multicolumn{5}{@{}l@{}}{\makebox[\linewidth][l]{\textbf{Nemotron-9B}\hfill{\scriptsize $S{=}16$, $T{=}8192$, 1.5\,req/s}}}\\
\cmidrule(l{0pt}r{0pt}){1-5}
Radix & 2.44 & 1.442 & 14,338 & 75.939 \\
Radix+Timer5 & 2.41 & 1.442 & 10,539 & 75.947 \\
Radix+Timer10 & 2.38 & 1.442 & 10,471 & 75.950 \\
Radix+Flex & 2.43 & 1.442 & 10,455 & 75.941 \\
\rowcolor{gray!12}
RadixFlow & 2.15 & 1.442 & \textbf{9,613} & 75.950 \\
RadixFlow+Frag & \textbf{2.11} & 1.442 & 9,861 & \textbf{75.959} \\
\cmidrule(l{0pt}r{0pt}){1-5}
\addlinespace[1pt]
\multicolumn{5}{@{}l@{}}{\makebox[\linewidth][l]{\textbf{Qwen3-8B (FA)}\hfill{\scriptsize $S{=}16$, $T{=}8192$, 5.0\,req/s}}}\\
\cmidrule(l{0pt}r{0pt}){1-5}
Radix & 16.4 & 4.000 & 9,223 & 85.243 \\
Radix+Timer5 & 16.7 & 3.996 & 6,779 & 85.209 \\
Radix+Timer10 & 16.4 & 4.001 & \textbf{6,536} & \textbf{85.265} \\
Radix+Flex & \textbf{16.3} & 4.002 & 6,782 & 85.250 \\
\rowcolor{gray!12}
RadixFlow & 16.4 & 4.000 & 6,746 & 85.250 \\
RadixFlow+Frag & 16.4 & 4.002 & 6,761 & 85.254 \\
\bottomrule
\end{tabular*}
\end{minipage}
\caption{RadixFlow lowers mean client latency on the hybrid models while preserving almost the same reused input. The first three methods are baselines; the last three are our ablation, main method, and extension. Each block covers 9306 requests across three regimes and two orders. Replay throughput divides completions by arm wall time, including arrivals and drain; it is not a saturation limit. Bold marks the best latency, round count, and reuse. FA denotes full attention.}
\label{tab:main}
\end{table*}

\begin{figure*}[t]
\centering\includegraphics[width=\textwidth]{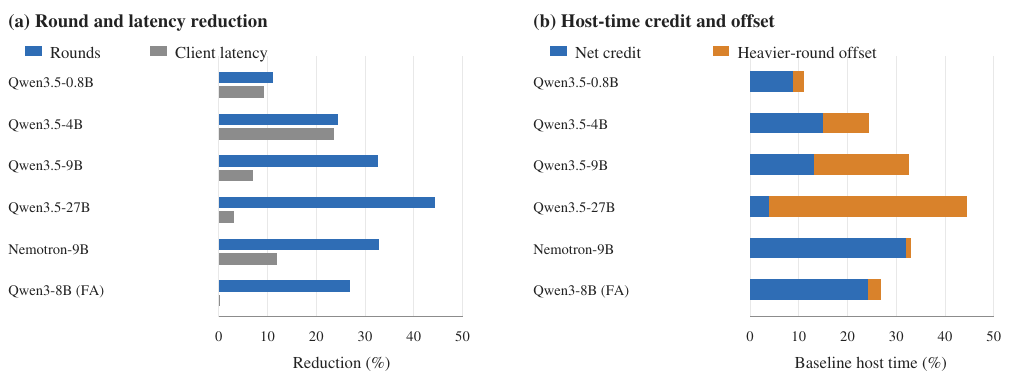}
\caption{Round savings and latency gains differ across the measured configurations. Left: reductions relative to Radix. Right: blue is net host-time credit and orange is the cost shift of remaining rounds; together they equal the gross round credit under Equation~\ref{eq:hostdecomposition}. The panels share model order.}
\label{fig:tradeoff}
\label{fig:hostdecomposition}
\end{figure*}

\begin{figure*}[t]
\centering\includegraphics[width=\textwidth]{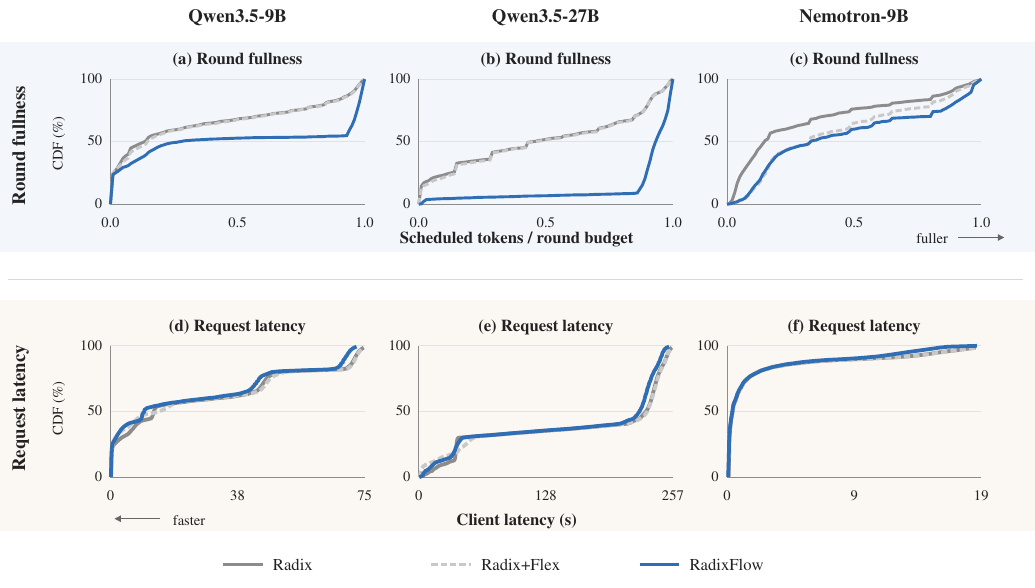}
\caption{(a)--(c) Round fullness: RadixFlow shifts rounds toward fuller budgets; rightward is better. (d)--(f) Request latency: the benefit reaches requests only partly and varies by model; leftward is faster. Curves pool three regimes and two orders per model. Latency is measured after submission and displayed through the largest p99 among the three policies.}
\label{fig:distributions}
\end{figure*}

\begin{table*}[t]
\centering\footnotesize\setlength{\tabcolsep}{5pt}
\begin{tabular*}{\textwidth}{@{\extracolsep{\fill}}lrrrrrr@{}}
\toprule
\textbf{Model} & \multicolumn{3}{c}{\textbf{Rows / round} $\uparrow$} & \multicolumn{3}{c}{\textbf{First selection (s)} $\downarrow$}\\
\cmidrule(lr){2-4}\cmidrule(lr){5-7}
 & Radix & RadixFlow & Gain (\%) & Radix & RadixFlow & Reduction (\%) \\
\midrule
Qwen3.5-0.8B & 3.63 & 4.10 & 12.9 & 0.057 & 0.049 & 14.0 \\
Qwen3.5-4B & 5.42 & 7.31 & 34.9 & 12.576 & 9.273 & 26.3 \\
Qwen3.5-9B & 6.61 & 10.02 & 51.6 & 25.682 & 23.255 & 9.5 \\
Qwen3.5-27B & 6.51 & 12.11 & 86.0 & 160.538 & 153.314 & 4.5 \\
Nemotron-9B & 2.91 & 4.38 & 50.5 & 1.898 & 1.582 & 16.6 \\
Qwen3-8B (FA) & 4.28 & 5.86 & 36.9 & 15.817 & 15.813 & 0.0 \\
\bottomrule
\end{tabular*}
\caption{Denser batches accompany earlier selection on hybrids. First selection measures registration-to-selection waiting; percentage changes use displayed values. See Appendix~\ref{app:outputaudit} for output sensitivity.}
\label{tab:coreaudit}
\end{table*}

\subsection{Setup}
\paragraph{Workload.}
We need heavy prefix sharing and little decode, so we build the workload from QuALITY and RACE reading comprehension \citep{pang2022quality,lai2017race}: 1034 QuALITY questions from 62 articles and 517 RACE-M questions from 128 articles. Each question becomes one logical request with four branches. Each branch holds the document, the question, and one candidate answer, and generates one token. The shared-document, candidate-specific structure models the prefill pattern of multi-candidate scoring and candidate-comparison judging. We measure serving performance rather than answer accuracy. Each model uses its own tokenizer, and overlength inputs are rejected rather than truncated.

Cold streams represent unrelated questions that share no cache across requests, while reuse streams represent repeated questions about a shared document. Mixed streams interleave documents. We implement these regimes with cache namespaces and fixed request orders; branches within a question still share its prefix. Three regimes and two input orders give 9306 logical requests per policy.

\paragraph{Backend and models.}
All policies run in vLLM 0.29 on A800 GPUs with eager execution, asynchronous scheduling disabled, and allocator utilization 0.8. They share the prefix index, the native state check, model weights, and cache primitives, so only the scheduling rule changes. We test four hybrid Qwen3.5 models (0.8B, 4B, 9B, and 27B), including Qwen3.5-0.8B-Base, shortened to Qwen3.5-0.8B. NVIDIA-Nemotron-Nano-9B-v2 is a Mamba2 hybrid, shortened to Nemotron-9B \citep{nvidia2025nano}. Qwen3-8B is the full-attention control, denoted Qwen3-8B (FA) \citep{qwen2026model,qwen2026hybrid}. 

\paragraph{Calibration.}
To avoid favoring any policy, we tune the row and token budgets once for the native scheduler on 64 held-out questions, following vLLM's automatic tuner \citep{vllm2026autotune}, and then freeze them. The chosen budgets appear in Table~\ref{tab:main}. All Qwen runs use an offered rate of five requests per second, which overloads the larger models. Nemotron-9B runs at 1.5 requests per second, which is 80\% of its calibrated native throughput. Appendix~\ref{app:protocol} gives the full protocol.

\paragraph{Policies.}
The six policies compare scheduling components on a common backend. Radix retains native vLLM admission and traversal within the shared experimental overlay. The overlay adds registration, indexing, tracing, and the common copy guard; Radix is not an untouched vLLM build. Radix+Timer5 and Radix+Timer10 add bounded waiting with two deadline settings \citep{triton2025batching}. Radix+Flex adds our admission rule. RadixFlow adds state-ready traversal on top, and RadixFlow+Frag adds the fragmentation guard. So Radix versus Radix+Flex tests admission, and Radix+Flex versus RadixFlow tests traversal.

\paragraph{Metrics.}
We report completed-replay throughput and post-submission latency, with sender lag separately tabulated in Appendix~\ref{app:savedtraces}. Rounds count nonempty global worker calls; reuse counts native reported hit tokens. Worker time is the elapsed host-side call interval (Appendix~\ref{app:accounting}).

\subsection{Q1: Serving performance}
RadixFlow reduces mean client latency by 9.3\%, 23.6\%, 7.1\%, and 3.1\% on the Qwen3.5 scale range, and by 12.0\% on Nemotron-9B (Table~\ref{tab:main}). Reused input stays within 0.03 percentage points of Radix on each hybrid. Completed-replay throughput rises by 4.2\%, 4.0\%, and 3.0\% on Qwen3.5-4B, 9B, and 27B. On the lighter Qwen3.5-0.8B and Nemotron runs, completion throughput tracks the offered arrivals while request latency falls. Each improvement compares policies under the same model, frozen budgets, and arrival trace.

\subsection{Q2: Admission and state-ready traversal}
The incremental comparison separates admission from traversal. On Qwen3.5-27B, Radix+Flex saves 1.6\% of the rounds, whereas RadixFlow saves 44\%. The difference persists in each stream: traversal adds a reduction of 44\%, 43\%, and 44\% relative to Radix+Flex in cold, mixed, and reuse streams. Here cold means no sharing across questions; the four branches of a question still share a prefix. A same-round dependency can therefore arise even in a cold stream.

Figure~\ref{fig:distributions}(a)--(c) and Table~\ref{tab:coreaudit} show denser rounds and earlier selection. On the full-attention control, Flex uses 6782 rounds and RadixFlow 6746. This small traversal increment is consistent with its recurrent-state check remaining inactive, while admission still groups work. Nemotron-9B gains chiefly from admission: it removes 27\% of rounds, followed by another 8\% from traversal. Its native occupancy of 2.91 rows out of 16 leaves room for admission to pack more work at the lower offered load.

\subsection{Q3: Round savings and their cost}
A denser round processes more work, so round count and elapsed time change by different amounts. Figure~\ref{fig:hostdecomposition} measures this conversion.

\paragraph{Worker-time accounting.}
\label{sec:cost}
Let $I_r=|x_r|$ and let $H_r$ be the cache-hit tokens that the engine reports for branch $r$. To measure how expensive each worker call is, we use its elapsed host-wrapper interval, which can include GPU waiting as well as CPU work. With this round cost $c(\mathcal B_t)$, we define
\begin{equation}
 \rho=\frac{\sum_r H_r}{\sum_r I_r},\quad T_{\mathrm{host}}=\sum_{t=1}^{K}c(\mathcal B_t),\quad \bar c=\frac{T_{\mathrm{host}}}{K}.
\label{eq:accounting}
\end{equation}
Token reuse $\rho$ measures how much input is skipped. $K$ and $\bar c$ measure how the rest is executed. For a baseline (index 0) and a policy (index 1), Equation~\ref{eq:accounting} gives the exact identity
\begin{equation}
 T_{\mathrm{host},0}-T_{\mathrm{host},1}
 =\underbrace{(K_0-K_1)\,\bar c_0}_{\text{round credit}}
 -\underbrace{K_1(\bar c_1-\bar c_0)}_{\text{cost shift}}.
 \label{eq:hostdecomposition}
\end{equation}
The identity separates saved invocations from the changed cost of the remaining work. We interpret it alongside client latency and completed-replay throughput. Packing replaces many small invocations with fewer larger ones. On Qwen3.5-27B, tokens per round rise by 80\%, but host cost per thousand scheduled tokens falls from 183 to 175\,ms. Total host-wrapper time falls by 3.8\%, alongside the 3.0\% replay-throughput gain. Figure~\ref{fig:hostdecomposition} separates round credit from the retained-round cost shift; on Qwen3.5-9B, that shift absorbs about 60\% of the credit.

\paragraph{Waiting and tail latency.}
Table~\ref{tab:coreaudit} shows earlier native selection on every hybrid. On Qwen3.5-9B, p95 falls from 73.3 to 70.6\,s; the full-attention control has nearly unchanged tails. Figure~\ref{fig:distributions}(d)--(f) complements the mean latency with the request distribution.

\paragraph{Selection evidence.}
Holding admission fixed, the Flex--RadixFlow comparison isolates the traversal increment. The higher occupancy and earlier selection in Table~\ref{tab:coreaudit} provide complementary evidence for improved batch construction. Appendix~\ref{app:savedtraces} specifies the trace scope, and Appendix~\ref{app:outputaudit} reports output sensitivity.

On Qwen3.5-27B, +Frag lowers latency by another 5.0\% relative to RadixFlow, with 3.6\% fewer rounds and 1.66 percentage points more reuse. Deferring a shortened aligned fragment changes both batch packing and the cache boundaries later consumers encounter, coupling fragmentation control with the realized cache trajectory.\footnote{Repeated scheduled tokens account for 0.14\% of Radix work and 0.39\% of RadixFlow work on 27B. They remain in all work and cost totals; Appendix~\ref{app:27accounting} gives the interval ledger.}

The hybrid ablations and full-attention control separate traversal from admission. Occupancy rises as rounds fall, and worker-time accounting quantifies how their cost affects completion. The observed request benefit is therefore interpreted through selection and execution cost, rather than through cache hits alone.

\paragraph{Arrival order and the operating point.}
Arrival order changes both the cache trajectory and the branches available for packing. RadixFlow lowers mean latency in both orders on every hybrid model (Appendix~\ref{app:savedtraces}). Offered load determines how those packing opportunities reach the client. With a small queue, earlier selection can shorten waiting even when completed throughput follows the arrival rate. Under sustained overload, packing instead helps drain accumulated work. Qwen3.5-27B illustrates the latter setting: its five-request/s offered rate exceeds its replay completion rate, so queueing dominates client latency. Throughput and total worker time quantify how efficiently this backlog is processed.

The full-attention control clarifies the role of state readiness. Admission can still consolidate branches that share a prefix, but traversal contributes little when the recurrent-state check is inactive. Hybrid models expose an additional opportunity: preserve the producer's work, block its consumer, and use remaining capacity for an independent branch.

\section{Conclusion}
In hybrid LLM serving, a cache match is not yet a runnable branch. RadixFlow blocks only the consumer of a same-round dependency and keeps selecting independent work that fits. Schedulers should therefore separate such dependencies from memory pressure: a blocked consumer needs only a recheck in the next round.

\section*{Limitations}
Our evaluation focuses on candidate-conditioned prefill with one generated token under frozen model, budget, and arrival settings. Worker-call timing and decoded-output agreement are the reported measures. Load sweeps, production scheduling, and finer dependency traces extend this setting; Appendix~\ref{app:baselineaudit} records the comparison scope.

\section*{Ethical Considerations}
Experiments use public benchmark inputs and publicly available model weights under their terms. Exact-prefix namespaces remain isolated, and no approximate semantic cache reuse is introduced. Our scripts keep failed and incomplete artifacts. Final summaries are produced only after identifier, ownership, and token-accounting checks. Appendix~\ref{app:pitfalls} documents the native state contract and shared implementation primitives. AI assistance was used for implementation and manuscript revision. The authors remain responsible for the reported evidence and claims.

\bibliography{custom}

\appendix
\section{Design Rationale and Alternatives}
\label{app:design}
This appendix explains the decisions that led to the reported method. The observations come from development prototypes and implementation audits. They are not additional benchmark results. Our final evaluation does not pool exploratory runs with the frozen comparisons.

\subsection{From prefix matches to executable work}
Our first question was whether a more precise index could turn more exact matches into useful cached computation. A radix index can locate a shared token boundary, but a hybrid model may lack the recurrent state at that boundary. Counting these matches as executable hits would overstate reuse. This led us to separate three events: finding a prefix, having a complete valid state, and placing the work into a legal worker round.

This separation changed the optimization target. Saving tokens is useful, but it does not show that a global round disappears. A shorter branch may still need the same number of rounds because of row limits, aligned chunks, the progress of other requests, or a dependency that resolves only after an existing round. We therefore kept cached tokens as an accounting quantity and added worker rounds as a scheduling quantity. Neither replaces elapsed time.

\subsection{Why not save a checkpoint at every branch?}
A checkpoint at a newly observed branch helps future reuse only if its complete state is available. Building it with an extra serial prefill may cost the time it is meant to save. Capturing it during an ordinary forward pass avoids that extra producer. It still costs state copies, publication, temporary buffers, and retention. A split in the token tree cannot recover a recurrent state that was never saved.

Our prototypes explored opportunistic capture and chained publication. They also revealed unused checkpoints and memory that stayed resident after a logical node became invalid. These observations led to a conservative choice. The reported method leaves checkpoint construction and eviction to the native backend and changes only the response to a recognized same-round dependency. State capture at arbitrary boundaries remains a separate design problem.

\subsection{Why not wait for every group?}
Waiting for a whole group can build a denser batch. It can also hold a cold request while its other branches arrive, even though no useful shared state is ready. Later revalidation protects correctness, but it cannot refund the waiting. A cache-aware gate solves this only if its readiness information stays valid and held requests keep normal cancellation and timeout behavior.

We therefore separate admission from state readiness. Flexible admission uses only arrived work and a short waiting bound, and releases work earlier when runnable work or resource demand justifies it. State-ready traversal responds to one specific native rejection rather than making all branches of a group wait together. The timer controls test whether ordinary bounded waiting explains the gain without the same-round dependency signal.

\subsection{Why not minimize rounds alone?}
A round with more rows or tokens can cost more than the rounds it replaces. A small change in packing can also move execution to a different kernel shape. In development, fewer tokens or rounds sometimes did not shorten completion time. This led us to keep round cost and client latency as separate measurements. The full-attention control matters here because an inactive recurrent rule should not receive credit for admission gains.

The same reasoning motivates the fragmentation extension. Deferring a branch can avoid a short aligned segment, but it can also delay useful progress. RadixFlow+Frag is therefore an ablation, not an assumed improvement. We do not pick the main method separately for each dataset after seeing which variant wins.

\section{Replay and Calibration Protocol}
\label{app:protocol}

\begin{figure*}[t]
\centering\includegraphics[width=\textwidth]{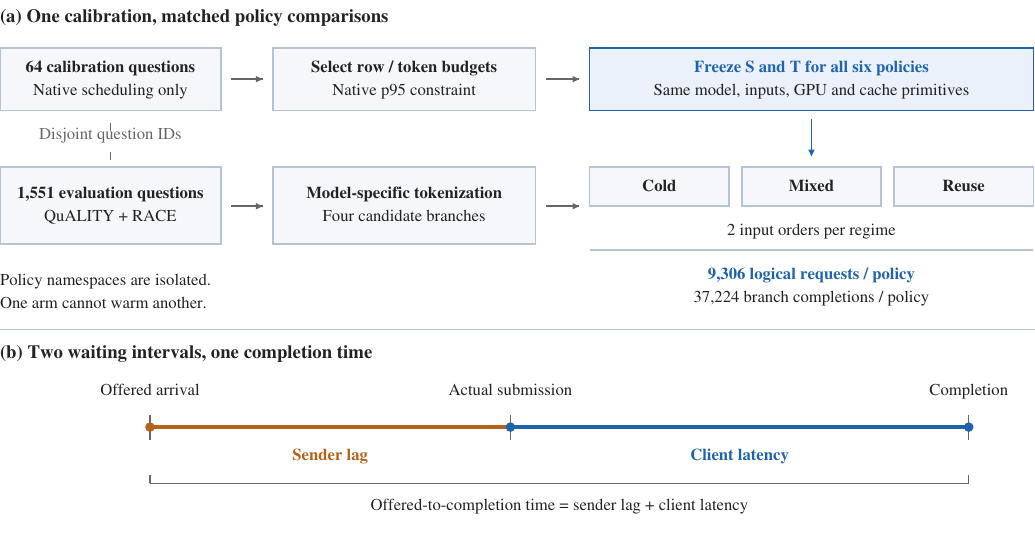}
\caption{Frozen budgets and matched inputs isolate the scheduling comparison within each model. (a) Native-only calibration fixes the row and token budgets before evaluation. Each policy receives the same 1551 questions in three regimes and two input orders. (b) Sender lag precedes actual submission; client latency starts at submission. The timeline is schematic, with no time scale.}
\label{fig:appprotocol}
\end{figure*}

Figure~\ref{fig:appprotocol} connects the frozen comparison protocol to the request timestamps used in the analysis.
\subsection{Input construction}
The main replay uses the same 1551 question identifiers for every policy of a model. A branch concatenates the source context, the question, one candidate answer, and a short correctness prompt. The four candidates produce four input encodings, and each generates one token. This construction keeps realistic shared documents and suffix differences while focusing measurement on prefill. It does not reproduce the full answer-scoring protocols of the benchmarks.

Each model's own tokenizer is applied before evaluation. We record the input identifier and token-sequence hash, and check positive lengths and the declared context limit. Token lengths can differ across models. No method receives labels, future arrivals, precomputed prefix cuts, or another policy's state. Overlength inputs are rejected rather than silently truncated or run with an unreported context extension.

\subsection{Cache isolation and arrival traces}
Cold requests receive question-specific cache namespaces. Shared-prefix conditions use the article namespace. Each policy run uses separate namespaces, so one arm cannot warm another. A logical request identifier is distinct from its cache namespace. Several concurrent requests can share an article without becoming one request group.

The Qwen series uses a common offered trace of five requests per second with forward and reversed assignments. Reversal changes the assignment order. It does not let any policy inspect future inputs. The client has a finite submission pool, so actual sends can lag offered timestamps under overload. We record offered time, actual send time, and completion time to expose this backpressure. A cache hit does not remove queueing, and prior serving studies examine load imbalance and tail latency directly \citep{sun2024llumnix,zhong2024distserve}.

\subsection{Native-only parameter selection}
Calibration uses 64 question identifiers disjoint from evaluation, under three reuse conditions. All candidate measurements use native scheduling. RadixFlow results are never used to choose the configuration. The selected configuration, precision, model revision, tokenizer, and cache geometry are then fixed across all six policies.

The search adapts the official tuning flow \citep{vllm2026autotune}. It uses real inputs instead of synthetic lengths and p95 instead of p99. We report this adaptation because a configuration tuned on a proxy workload can behave differently on long shared-prefix requests. A configuration that meets the calibration constraint may still exceed the same p95 on the longer replay.

\subsection{Pairing and exclusions}
Within a model, all policies use the same GPU and workload definition. Across models, services may share host resources. A foreign GPU process entering a private experiment invalidates that attempt for timing comparisons. Only experiment-owned processes are stopped, and partial outputs are kept separately.

Completed arms are checked for identifiers and expected counts before aggregation. A failed or partial arm is never filled in from another run. Repeated traffic conditions are not additional independent questions.

\section{Accounting and Timing Semantics}
\label{app:accounting}
\subsection{Tokens and global rounds}
For branch $r$, let $I_r$ be its prompt tokens and $H_r$ its native reported cache-hit tokens, as in Equation~\ref{eq:accounting}. We report
\[
 \rho=\frac{\sum_r H_r}{\sum_r I_r},\qquad
 W=\sum_r(I_r-H_r)+E.
\]
Here $E$ is additional scheduled work identified from per-request intervals. It is zero in the earlier validated runs and is kept explicitly in the 27B ledger.

A worker round is one global worker invocation with scheduled model work. Empty executor calls for completion or cleanup are counted separately. Summing per-request visits or tensor-parallel rank calls would overcount global rounds. Cleanup calls are not deleted merely because they carry zero tokens, since they may have lifecycle duties. For a fixed row capacity, the unweighted bubble count is $\sum_t(S-|\mathcal B_t|)$. It counts slots rather than tokens or hardware use, so we use it only for the illustration in Figure~\ref{fig:mechanism}.

\subsection{Three distinct clocks}
Client latency measures the completion of the required branches after actual submission. Offered-to-completion time also captures sender lateness. Host-wrapper time measures a CPU-side call around worker execution, and asynchronous returns and enqueue gaps can separate it from GPU work. CUDA event intervals, where available, measure stream intervals and may include launch gaps.

These clocks are not interchangeable. CUDA timing guidance distinguishes asynchronous execution from properly ordered timing \citep{nvidia2026timing}. We never add a CPU enqueue duration to a CUDA interval as if the two were disjoint. A drop in host-wrapper time is not a direct estimate of saved GPU time, especially when batch shapes or queue occupancy change.

\subsection{First-token edge cases}
A generated token may decode to an empty string or a special symbol. A successful one-token response can therefore lack a first nonempty text event. Such a response is neither a timeout nor a zero-latency observation. Missing text-event times stay missing, and conditional TTFT summaries report their coverage. Client completion latency and token usage remain available and are checked independently.

\section{Full Stream Model and Traversal Properties}
\label{app:formalmodel}
Let $\mathcal V$ be the token vocabulary and $[n]=\{1,\ldots,n\}$.
The four-round layout is as short as capacity allows. The following bound makes this precise.

\begin{proposition}[Capacity bound]
\label{prop:bound}
Fix a hit assignment and a partition of the remaining work into fragments. Let $W$ be their total tokens, $V$ their total row visits, and $D$ the length of their longest dependency chain in rounds. Every schedule $\pi$ that realizes them satisfies
\begin{equation}
 K(\pi)\geq\max\!\left\{\left\lceil W/T\right\rceil,\left\lceil V/S\right\rceil,D\right\}.
\label{eq:lowerbound}
\end{equation}
\end{proposition}
For Figure~\ref{fig:mechanism}, $T=4096$ and $S=4$, so $\lceil12{,}970/T\rceil=4$ and $\lceil15/S\rceil=4$. The longest illustrated dependency chain has two fragments. The feasible four-round layout therefore attains the bound for this fixed work. The proof is in Appendix~\ref{app:proofs}; general online streams also face arrival constraints.

\begin{definition}[State readiness]
\label{def:ready}
A model fixes a set $\Sigma$ of state components. Full attention has $\Sigma=\{\mathrm{kv}\}$, and a hybrid model has $\Sigma=\{\mathrm{kv},\mathrm{rec},\mathrm{conv}\}$. Let $\mathcal C_t$ be the set of pairs $(p,\varsigma)$ such that component $\varsigma\in\Sigma$ of prefix $p$ is published and passes the native state check at the start of round $t$. A boundary $h$ is \emph{ready} for branch $r$ in round $t$ if $(x_{r,\le h},\varsigma)\in\mathcal C_t$ for all $\varsigma\in\Sigma$. The \emph{ready length} $h_t(r)$ is the largest ready boundary, and $h_t(r)=0$ means cold execution.
\end{definition}
The prefix index reports the longest token match $\hat h_t(r)\ge h_t(r)$. For full attention, the KV pages of a matched prefix are usually present, so $\hat h_t(r)=h_t(r)$. In a hybrid model, $\hat h_t(r)>h_t(r)$ whenever the recurrent component at the matched boundary is missing. State produced in round $t$ enters $\mathcal C_{t+1}$, not $\mathcal C_t$. For $A$, the gap between matched length $\hat h_t(A)$ and ready length $h_t(A)$ separates prefix identity from published state. It creates an avoidable bubble when the scan stops despite other feasible work.

\begin{definition}[Prefill request stream]
\label{def:stream}
A stream is a finite sequence $\mathcal Q=((q_i,a_i))_{i\in[n]}$ with offered arrival times $a_1\le\cdots\le a_n$. Request $q_i=(x_{i1},\ldots,x_{im_i})$ holds $m_i$ input branches $x_{ij}\in\mathcal V^{*}$. Let $\mathcal R=\{(i,j):i\in[n],j\in[m_i]\}$ be the set of branches. For $r=(i,j)$ we write $x_r=x_{ij}$ and $a_r=a_i$. An online policy observes branch $r$ only after time $a_r$.
\end{definition}
A branch is one input encoding. In Figure~\ref{fig:mechanism}, $P$ and $A$ are branches with a shared prefix; our replay groups four branches into each logical request.

\begin{definition}[Feasible batch and schedule]
\label{def:feasible}
Let $\mathcal A_t\subseteq\mathcal R$ be the branches that have arrived, have been released by admission by round $t$, and still have unscheduled tokens. A batch $\mathcal B_t\subseteq\mathcal A_t$ assigns each $r\in\mathcal B_t$ a token count $n_t(r)\ge 1$. It is \emph{feasible} if (i) $|\mathcal B_t|\le S$, (ii) $\sum_{r\in\mathcal B_t}n_t(r)\le T$, and (iii) every $r\in\mathcal B_t$ resumes from a boundary that is ready in round $t$ or from its own state of an earlier round. A schedule $\pi=(\mathcal B_1,\ldots,\mathcal B_K)$ is a sequence of nonempty feasible batches that covers every token of every branch. Its round count is $K(\pi)$.
\end{definition}
Condition (iii) prevents $A$ from using state that $P$ is still producing. An independent branch such as $D$ can join $P$ if the row and token budgets allow it.

\paragraph{Properties.}
Within a round, $\mathcal C_t$ is fixed and the residual budgets only shrink as $\mathcal B$ grows. Adding a branch can also create a new producer. The evaluated synchronous scheduler publishes no completed worker state during this scan, and its current-step rejection remains active until a later step. Thus a blocked $A$ does not become selectable as more branches join the partial batch. We state the corresponding monotonicity requirement explicitly below; additional backend outcomes must satisfy it for part (iii) of the proposition.

\begin{assumption}[Monotone readiness]
\label{ass:monotone}
For any round $t$ and partial batches $\mathcal B\subseteq\mathcal B'$, if $\sigma_t(r\mid\mathcal B')=\Ready{}$, then $\sigma_t(r\mid\mathcal B)=\Ready{}$.
\end{assumption}

\begin{proposition}[Properties of state-ready traversal]
\label{prop:traversal}
Fix round $t$, the state $\mathcal C_t$, the eligible set, and the order $\prec$. Let $\mathcal B^{\mathrm{stop}}_t$ and $\mathcal B^{\mathrm{RF}}_t$ be the batches built by prefix-stop and by RadixFlow. Then:
\begin{enumerate}
\setlength{\itemsep}{0pt}
\item[(i)] $\mathcal B^{\mathrm{RF}}_t$ is feasible.
\item[(ii)] $\mathcal B^{\mathrm{stop}}_t\subseteq\mathcal B^{\mathrm{RF}}_t$, hence $b^{\mathrm{RF}}_t\le b^{\mathrm{stop}}_t$.
\item[(iii)] Under Assumption~\ref{ass:monotone}, if no \Other{} outcome ends the scan, round $t$ has no avoidable bubble under RadixFlow.
\end{enumerate}
\end{proposition}
The proof is in Appendix~\ref{app:proofs}. Part (i) is the safety guarantee: blocking a consumer changes selection without accepting state rejected by the native state check. Parts (ii) and (iii) characterize the bubbles recovered by continuing past \Blocked{} outcomes, under the stated conditions. The comparison holds for a fixed state. Over a stream, the two policies reach different states, so the total round count is measured in Section~\ref{sec:setup}.

\section{Proofs}
\label{app:proofs}
\paragraph{Proof of Proposition~\ref{prop:bound}.}
Each round schedules at most $T$ tokens, so $K(\pi)\,T\ge W$ and $K(\pi)\ge\lceil W/T\rceil$. Each row visit occupies one of at most $S$ rows in one round, so $K(\pi)\ge\lceil V/S\rceil$. Consider two consecutive fragments on a dependency chain. The later one resumes from state that the earlier one produces. By Definition~\ref{def:ready}, that state enters $\mathcal C_{t+1}$ when the earlier fragment runs in round $t$. Condition (iii) of Definition~\ref{def:feasible} then places the later fragment in round $t+1$ or after. A chain of $D$ fragments thus needs $D$ distinct rounds, so $K(\pi)\ge D$. \hfill$\square$

\paragraph{Proof of Proposition~\ref{prop:traversal}.}
(i) RadixFlow starts from $\mathcal B=\emptyset$, which is feasible. Algorithm~\ref{alg:round} adds a branch only at line~\ref{alg:accept}; line~\ref{alg:defer} changes the wait queue without adding work. It adds $r$ only when $\sigma_t(r\mid\mathcal B)=\Ready{}$, which by Definition~\ref{def:outcome} means that $\mathcal B\cup\{r\}$ is feasible. A \Blocked{} branch is never added, and the native action for \Other{} never adds a branch that fails the native state check. By induction over the scan, every partial batch is feasible, and so is $\mathcal B^{\mathrm{RF}}_t$.

(ii) Let $r^\star$ be the first branch in $\prec$ whose outcome under prefix-stop is not \Ready{}. Before $r^\star$, both scans see the same partial batch at every step, obtain the same outcomes, and receive the same token counts from the native allocator. They therefore hold the same partial batch $\mathcal B^\star$ when they reach $r^\star$, and prefix-stop returns $\mathcal B^{\mathrm{stop}}_t=\mathcal B^\star$. If the outcome of $r^\star$ is \Other{}, RadixFlow takes the same native action, and the two batches coincide. If it is \Blocked{}, RadixFlow continues and only adds branches, so $\mathcal B^\star\subseteq\mathcal B^{\mathrm{RF}}_t$. If no such $r^\star$ exists, both scans select every eligible branch. In all cases $\mathcal B^{\mathrm{stop}}_t\subseteq\mathcal B^{\mathrm{RF}}_t$, and $b_t=S-|\mathcal B_t|$ gives $b^{\mathrm{RF}}_t\le b^{\mathrm{stop}}_t$.

(iii) Assume no \Other{} outcome ends the scan, so RadixFlow visits every branch of $\mathcal A_t$. Take any $r\in\mathcal A_t\setminus\mathcal B^{\mathrm{RF}}_t$, and let $\mathcal B_r$ be the partial batch when $r$ was visited. Since $r$ was not added, $\sigma_t(r\mid\mathcal B_r)\ne\Ready{}$. Partial batches only grow, so $\mathcal B_r\subseteq\mathcal B^{\mathrm{RF}}_t$. By Assumption~\ref{ass:monotone}, $\sigma_t(r\mid\mathcal B^{\mathrm{RF}}_t)\ne\Ready{}$. No branch outside the batch is \Ready{}, so by Definition~\ref{def:bubble} the round has no avoidable bubble. \hfill$\square$

\section{Native State Contract}
\label{app:pitfalls}
\subsection{Native prefix-stop path}
\label{app:baselinepath}
The reference path is vLLM 0.29.0, not a synthetic queue policy. In \texttt{Scheduler.schedule}, lines 991--1011 of \texttt{vllm/v1/core/sched/scheduler.py} allocate waiting-request blocks and end the waiting-queue scan when allocation fails \citep{vllm2026scheduler}. In \texttt{MambaManager.get\_num\_blocks\_to\_allocate}, lines 1476--1485 reject reuse of a block hash recorded in \texttt{cached\_blocks\_this\_step} by reporting a requirement above pool capacity \citep{vllm2026cache}. The same set is cleared at the next step in lines 1729--1730. This is the specific same-round dependency RadixFlow recognizes. Other native paths can already skip work for different reasons, and remain unchanged. All policies share eager execution, disabled asynchronous scheduling, prefix caching, and native state checks.

All policies share request registration, cache namespace isolation, component-generation checks, and the same verified copy primitive. The state check distinguishes published components from a producer's in-flight state; a blocked consumer stays registered and is woken for a later native recheck.

\section{Anonymous Reproducibility Record}
\label{app:anonymous}
The review manuscript uses the public model names, benchmark sources, and hardware descriptions needed to interpret the experiments. It omits author identities, affiliations, private repository URLs, hostnames, account names, absolute paths, and internal experiment identifiers. Development failures are described by mechanism rather than by chronology.

An anonymized release will include pinned model and data revisions, input and source hashes, frozen configurations, policy definitions, offered traces, per-arm identifier counts, worker and cleanup-call counts, and token-accounting summaries. Logs will be scrubbed of credentials and machine-specific paths before release.

\section{Timing and Extension Protocol}
\label{app:componentcontract}
\label{app:pending}
Component profiling aligns request IDs with worker rounds and separates CPU lookup, selection, dispatch, GPU restoration, prefill, and publication. Exclusive intervals preserve overlap rather than double-counting stages. Load sweeps at matched fractions of native capacity, fixed-row scale comparisons, and wider branching are natural extensions of the reported frozen replay.

\section{27B Token-Accounting Ledger}
\label{app:27accounting}
All six 27B policies completed 9306 logical requests each. The initial summary rejected the identity $W=\sum_r(I_r-H_r)$ because it omitted scheduled revisits. We kept the original outputs and rebuilt the interval ledger from actual executor-call traces.

Each request's first recorded start equals its reported cache-hit length. Its intervals stay within the prompt and reach the final prompt position without gaps. The union of intervals measures uniquely scheduled input positions. Overlapping intervals measure repeated scheduled work. Two Radix+Timer10 branches first start after a cached prefix and then return to position zero. Their reset work accounts for the remaining discrepancy. For every request and every arm, we verify
\[
 W=\sum_r(I_r-H_r)+E_{\mathrm{overlap}}+E_{\mathrm{reset}}.
\]
\begin{table}[t]
\centering\small
\begin{tabular}{lrrr}
\toprule
Policy & Overlap & Reset & Total extra\\
\midrule
Radix & 66,944 & 0 & 66,944\\
Radix+Timer5 & 49,584 & 0 & 49,584\\
Radix+Timer10 & 53,760 & 512 & 54,272\\
Radix+Flex & 48,864 & 0 & 48,864\\
RadixFlow & 182,384 & 0 & 182,384\\
RadixFlow+Frag & 186,096 & 0 & 186,096\\
\bottomrule
\end{tabular}
\caption{Additional scheduled tokens in the 27B replay. Worker calls and token totals are kept rather than deduplicated.}
\label{tab:27ledger}
\end{table}
This ledger reconciles reported hits with actual scheduled intervals (Table~\ref{tab:27ledger}). RadixFlow's overlap accounts for 0.39\% of its scheduled tokens, versus 0.14\% for Radix. Thus reported prefix reuse and additional scheduled work remain separate quantities. The traces identify repeated intervals and the two reset prefixes; attributing their cause to preemption would require native preemption events. Native reported token reuse, uniquely skipped input positions, and repeated scheduling are recorded separately. Identifier coverage, one-token completions, interval continuity, and GPU ownership checks pass for all arms.

\section{Saved-Trace Mechanism and Cost Analysis}
\label{app:savedtraces}

\begin{figure*}[t]
\centering\includegraphics[width=\textwidth]{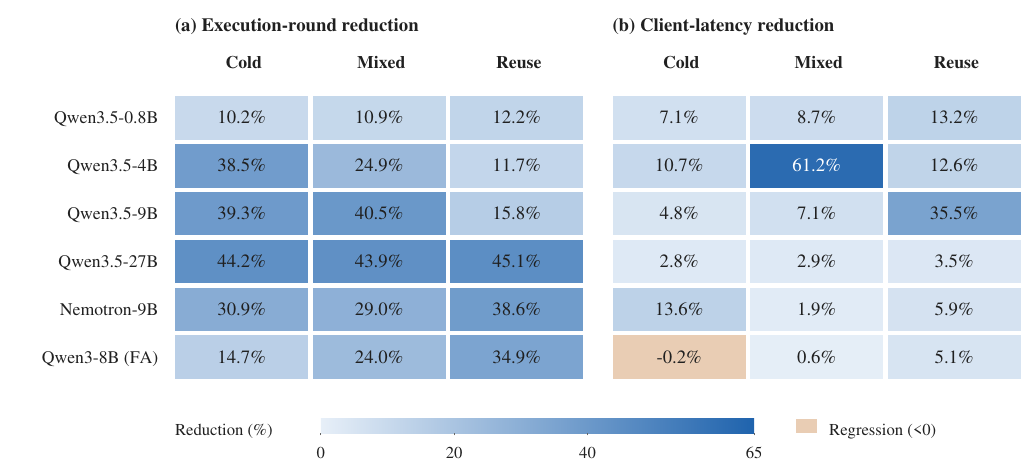}
\caption{Round reductions and latency gains respond differently to the reuse regime. Each cell compares RadixFlow with Radix on the same model and stream, pooling both input orders. Blue indicates a reduction; orange indicates a regression. The panels use the same percentage scale; cold prevents sharing across questions, not within one question. Values summarize the frozen replays in Table~\ref{tab:regimetraces}.}
\label{fig:appregimes}
\end{figure*}
We reanalyze the completed runs without additional model inference. Each worker record is joined to measured requests by branch identifier. Functionality smoke requests are excluded. We verify the row and token budgets for every joined round. Logical requests are counted once per policy, regime, and input order. Table~\ref{tab:regimetraces} separates admission from traversal; Table~\ref{tab:tails} reports all policies' latency distributions.
Figure~\ref{fig:appregimes} makes the regime dependence visible: the largest latency gain need not coincide with the largest round reduction.

Percentiles in this appendix use linear interpolation of the empirical order statistics. Client latency is submission to completion of the four branches. Sender lag is offered arrival to submission. Offered-arrival latency is their sum for each request, before computing a percentile. Percentiles of the two parts are not additive.

\begin{table*}[t]
\centering\small\setlength{\tabcolsep}{5pt}
\begin{tabular}{llrrr rr}
\toprule
Model & Stream & Radix $K$ & Flex $K$ & RadixFlow $K$ & Latency reduction (\%) & $K/L_{\mathrm{cap}}$\\
\midrule
Qwen3.5-0.8B & Cold & 6,849 & 6,563 & 6,152 & 7.1 & 3.33 \\
 & Mixed & 6,603 & 6,101 & 5,884 & 8.7 & 3.94 \\
 & Reuse & 6,351 & 5,745 & 5,574 & 13.2 & 3.88 \\
\addlinespace[2pt]
Qwen3.5-4B & Cold & 4,477 & 4,382 & 2,753 & 10.7 & 1.49 \\
 & Mixed & 3,866 & 3,805 & 2,905 & 61.2 & 2.56 \\
 & Reuse & 5,081 & 4,755 & 4,488 & 12.6 & 7.71 \\
\addlinespace[2pt]
Qwen3.5-9B & Cold & 4,389 & 4,318 & 2,666 & 4.8 & 1.44 \\
 & Mixed & 3,336 & 3,192 & 1,984 & 7.1 & 1.68 \\
 & Reuse & 3,317 & 3,260 & 2,794 & 35.5 & 4.18 \\
\addlinespace[2pt]
Qwen3.5-27B & Cold & 4,044 & 3,998 & 2,256 & 2.8 & 1.13 \\
 & Mixed & 3,874 & 3,820 & 2,172 & 2.9 & 1.14 \\
 & Reuse & 3,830 & 3,737 & 2,101 & 3.5 & 1.14 \\
\addlinespace[2pt]
Nemotron-9B & Cold & 4,821 & 4,045 & 3,332 & 13.6 & 1.59 \\
 & Mixed & 4,576 & 3,360 & 3,248 & 1.9 & 2.37 \\
 & Reuse & 4,941 & 3,050 & 3,033 & 5.9 & 3.47 \\
\addlinespace[2pt]
Qwen3-8B (FA) & Cold & 2,449 & 2,091 & 2,090 & -0.2 & 1.39 \\
 & Mixed & 2,268 & 1,732 & 1,723 & 0.6 & 1.89 \\
 & Reuse & 4,506 & 2,959 & 2,933 & 5.1 & 3.76 \\
\addlinespace[2pt]
\bottomrule
\end{tabular}
\caption{Regime analysis, pooling both orders. Latency reduction compares RadixFlow with Radix. Flex denotes Radix+Flex. $L_{\mathrm{cap}}$ sums the separate capacity bounds for the two orders using RadixFlow's observed work. Cold prevents sharing across questions, not within a question.}
\label{tab:regimetraces}
\end{table*}

\begin{table*}[t]
\centering\footnotesize\setlength{\tabcolsep}{6pt}
\begin{tabular}{llrrrrr}
\toprule
Model & Policy & p50 (s) & p95 (s) & p99 (s) & Sender lag (s) & Offered latency (s)\\
\midrule
Qwen3.5-0.8B & Radix & 0.217 & 0.503 & 0.767 & 0.000 & 0.256 \\
 & Radix+Timer5 & 0.225 & 0.512 & 0.842 & 0.000 & 0.261 \\
 & Radix+Timer10 & 0.227 & 0.526 & 0.828 & 0.000 & 0.265 \\
 & Radix+Flex & 0.210 & 0.521 & 0.798 & 0.000 & 0.249 \\
 & RadixFlow & 0.201 & 0.455 & 0.630 & 0.000 & 0.232 \\
 & RadixFlow+Frag & 0.200 & 0.443 & 0.606 & 0.000 & 0.228 \\
\addlinespace[2pt]
Qwen3.5-4B & Radix & 4.525 & 47.300 & 48.576 & 15.943 & 29.067 \\
 & Radix+Timer5 & 0.901 & 96.417 & 150.829 & 14.059 & 25.819 \\
 & Radix+Timer10 & 0.900 & 93.226 & 151.225 & 14.008 & 25.519 \\
 & Radix+Flex & 2.656 & 47.658 & 48.840 & 16.205 & 28.996 \\
 & RadixFlow & 1.666 & 43.718 & 44.928 & 11.875 & 21.907 \\
 & RadixFlow+Frag & 1.274 & 45.873 & 47.405 & 13.131 & 23.265 \\
\addlinespace[2pt]
Qwen3.5-9B & Radix & 12.849 & 73.326 & 75.169 & 60.581 & 87.248 \\
 & Radix+Timer5 & 2.316 & 155.450 & 254.673 & 60.875 & 87.404 \\
 & Radix+Timer10 & 2.290 & 158.283 & 251.954 & 59.667 & 86.191 \\
 & Radix+Flex & 14.205 & 73.509 & 75.208 & 59.995 & 86.461 \\
 & RadixFlow & 9.978 & 70.605 & 72.574 & 54.651 & 79.425 \\
 & RadixFlow+Frag & 10.045 & 72.476 & 74.510 & 55.190 & 80.047 \\
\addlinespace[2pt]
Qwen3.5-27B & Radix & 229.697 & 252.283 & 256.890 & 721.476 & 885.663 \\
 & Radix+Timer5 & 11.554 & 769.029 & 970.131 & 742.837 & 911.084 \\
 & Radix+Timer10 & 11.273 & 911.466 & 1014.259 & 743.882 & 912.454 \\
 & Radix+Flex & 230.947 & 252.283 & 256.511 & 717.981 & 881.401 \\
 & RadixFlow & 223.046 & 246.640 & 251.870 & 696.686 & 855.808 \\
 & RadixFlow+Frag & 199.937 & 253.765 & 262.483 & 653.534 & 804.627 \\
\addlinespace[2pt]
Nemotron-9B & Radix & 0.420 & 15.818 & 18.789 & 0.000 & 2.438 \\
 & Radix+Timer5 & 0.393 & 12.128 & 40.586 & 0.000 & 2.407 \\
 & Radix+Timer10 & 0.396 & 11.799 & 39.126 & 0.000 & 2.378 \\
 & Radix+Flex & 0.395 & 15.962 & 18.916 & 0.000 & 2.432 \\
 & RadixFlow & 0.391 & 13.275 & 16.133 & 0.000 & 2.146 \\
 & RadixFlow+Frag & 0.392 & 13.492 & 16.949 & 0.000 & 2.111 \\
\addlinespace[2pt]
Qwen3-8B (FA) & Radix & 3.235 & 54.603 & 55.425 & 24.775 & 41.198 \\
 & Radix+Timer5 & 1.754 & 94.325 & 117.654 & 25.229 & 41.910 \\
 & Radix+Timer10 & 1.657 & 97.192 & 104.945 & 24.484 & 40.843 \\
 & Radix+Flex & 3.594 & 54.596 & 55.342 & 24.743 & 41.086 \\
 & RadixFlow & 3.631 & 54.638 & 55.402 & 24.868 & 41.278 \\
 & RadixFlow+Frag & 3.579 & 54.588 & 55.420 & 24.882 & 41.291 \\
\addlinespace[2pt]
\bottomrule
\end{tabular}
\caption{Latency distribution and mean sender lag on all measured requests. The last two columns are means, not tail quantiles. The offered rate differs between Qwen and Nemotron-9B.}
\label{tab:tails}
\end{table*}

\subsection{Batch shape and cost association}
We fit $t_k=\alpha+\beta(n_k/1000)+\gamma v_k+\delta\mathbf{1}[\mathrm{RadixFlow}]$ to nonempty rounds from Radix, Radix+Flex, and RadixFlow. Here $n_k$ is scheduled tokens and $v_k$ is row visits. Table~\ref{tab:shape} reports the fit alongside mean batch shape. Cache lengths, copied bytes, kernel shapes, and temporal host load are omitted variables. The adjusted coefficient $\delta$ summarizes policy association under this batch-shape model; component profiling separates its underlying costs.

\begin{table*}[t]
\centering\small
\begin{tabular}{lrrrrrr}
\toprule
Model & \multicolumn{2}{c}{Tokens/round} & \multicolumn{2}{c}{Rows/round} & $\delta$ (ms) & $R^2$\\
 & Radix & RadixFlow & Radix & RadixFlow & & \\
\midrule
Qwen3.5-0.8B & 1401 & 1575 & 3.63 & 4.10 & 0.51 & 0.328 \\
Qwen3.5-4B & 2173 & 2871 & 5.42 & 7.31 & 2.70 & 0.911 \\
Qwen3.5-9B & 2739 & 4060 & 6.61 & 10.02 & 3.05 & 0.957 \\
Qwen3.5-27B & 3998 & 7208 & 6.51 & 12.11 & 18.41 & 0.997 \\
Nemotron-9B & 2482 & 3701 & 2.91 & 4.38 & -0.61 & 0.238 \\
Qwen3-8B (FA) & 2328 & 3181 & 4.28 & 5.86 & 0.12 & 0.232 \\
\bottomrule
\end{tabular}
\caption{Batch-shape changes and descriptive host-cost regression. Separate fits are used for each model.}
\label{tab:shape}
\end{table*}

\begin{figure*}[t]
\centering\includegraphics[width=\textwidth]{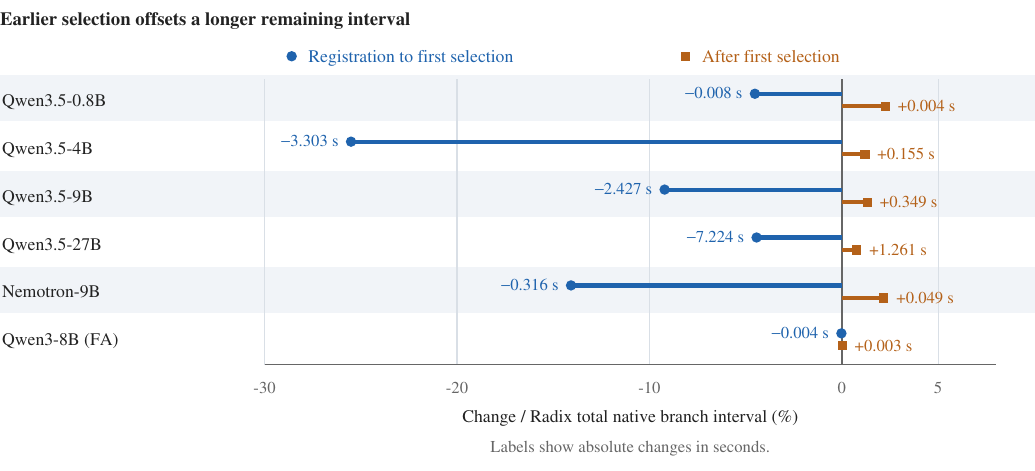}
\caption{Earlier selection offsets a longer interval after selection on the hybrid models. Points show the RadixFlow-minus-Radix change in each interval, normalized by the sum of the two Radix mean intervals. Labels give absolute changes in seconds, calculated from the rounded means in Table~\ref{tab:nativewaiting}. The later interval includes intervening waiting and completion processing as well as execution; these are native branch intervals, not logical client latency.}
\label{fig:appwaiting}
\end{figure*}

\begin{table*}[t]
\centering\small
\begin{tabular}{lrrrr}
\toprule
Model & \multicolumn{2}{c}{Registration to first selection (s)} & \multicolumn{2}{c}{After first selection (s)}\\
 & Radix & RadixFlow & Radix & RadixFlow\\
\midrule
Qwen3.5-0.8B & 0.057 & 0.049 & 0.120 & 0.124 \\
Qwen3.5-4B & 12.576 & 9.273 & 0.373 & 0.528 \\
Qwen3.5-9B & 25.682 & 23.255 & 0.664 & 1.013 \\
Qwen3.5-27B & 160.538 & 153.314 & 2.457 & 3.718 \\
Nemotron-9B & 1.898 & 1.582 & 0.347 & 0.396 \\
Qwen3-8B (FA) & 15.817 & 15.813 & 0.322 & 0.325 \\
\bottomrule
\end{tabular}
\caption{Mean native branch intervals on all 37,224 measured branches per policy. First selection is the first scheduler selection, not GPU execution start. The second interval includes subsequent rounds, intervening waiting, and completion processing. These native branch intervals are not directly additive to logical client latency.}
\label{tab:nativewaiting}
\end{table*}
The native timestamps provide a second view of waiting (Table~\ref{tab:nativewaiting}). Figure~\ref{fig:appwaiting} places the two interval changes on a common within-model scale. On Qwen3.5-27B, mean registration-to-first-selection time falls by 7.22\,s, while the remaining interval grows by 1.26\,s. On Qwen3.5-9B, the corresponding changes are a 2.43\,s reduction and a 0.35\,s increase. This supports a concrete trade-off: requests enter execution sooner, but denser rounds extend their subsequent service interval. The latter includes waiting as well as execution and cannot isolate restoration cost.

\subsection{Capacity gap and counter scope}
For a fixed policy, stream, and order, $L_{\mathrm{cap}}=\max(\lceil W/T\rceil,\lceil V/S\rceil)$ relaxes arrivals, alignment, and state dependencies. $W$ includes repeated scheduled tokens and $V$ includes every scheduled fragment. These quantities can change with the policy, so this diagnostic is not a shared global optimum. The bound becomes loose when highly reused short fragments cannot arrive or become ready together. Likewise, unused row slots and token budget measure residual capacity, not automatically avoidable bubbles or GPU idle time.

The saved state-check counters are cumulative visits, including smoke traffic and repeated scans. They establish that the recurrent check is active on hybrid models and absent on the full-attention control. They do not identify the blocked and recovered rows in each worker round. Per-request blocking counts are not recorded. A future trace must capture the dependency, the skip, and the recovered row in the same batch before attributing a specific saved round to a specific bypass or analyzing starvation.

\subsection{Scale and load sensitivity}
A plausible scale effect is that longer worker calls let more requests accumulate, exposing more consumers of current-round producers. However, row capacity also changes from 16 to 64 and the offered load queues larger models more heavily. Per-round dependency and recovered-row traces would distinguish these factors.

\subsection{Dataset and order sensitivity}
\begin{figure*}[t]
\centering\includegraphics[width=\textwidth]{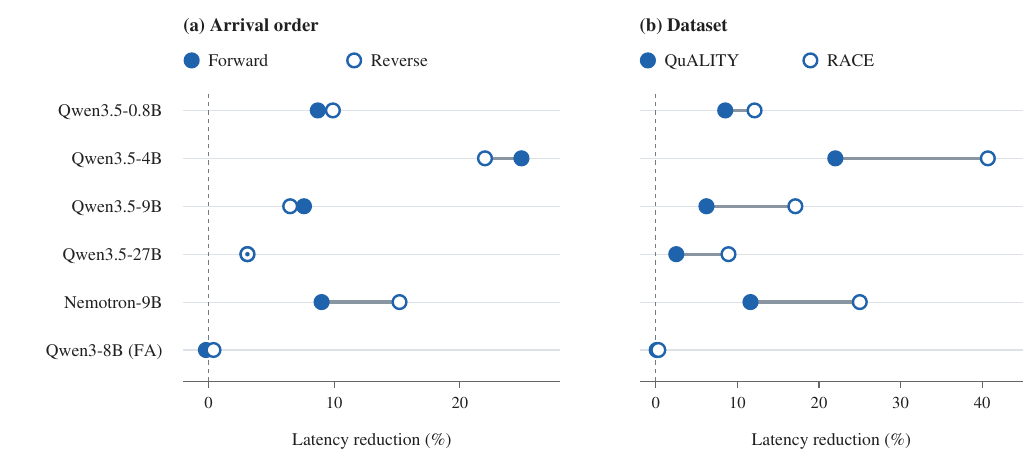}
\caption{Latency improves in both input orders and both datasets on every hybrid model. Points show the percentage reduction in mean client latency for RadixFlow relative to Radix, pooling the other conditions. Connectors pair the two strata within a model; they are not uncertainty intervals. Forward and reverse are different input assignments within the frozen evaluation.}
\label{fig:appstrata}
\end{figure*}

The workload uses article-level sharing, so arrival order changes both realized reuse and the executable set. We retain both orders in every primary comparison rather than selecting the faster one. Dataset-level latency summaries can be obtained by request identifiers, but global rounds cannot be assigned uniquely to QuALITY or RACE when a batch contains both. We therefore report global round counts at the stream level, where batch membership is unambiguous. Figure~\ref{fig:appstrata} shows the latency reduction in each stratum. Both orders favor RadixFlow on each hybrid model. The full-attention control changes sign across orders, with small effects in both datasets, consistent with its small aggregate difference.

\section{Extended Related Work}
\label{app:extendedrelated}
Serving research organizes four decisions: what state to keep, where to place requests, when to execute them, and how to compute a batch. Our focus is eligibility within batch construction: which matched branches can run now?

\paragraph{Keeping reusable state.}
Prompt Cache reuses predefined prompt modules \citep{gim2024promptcache}, while Pensieve preserves conversation state across memory tiers \citep{yu2023pensieve}. CachedAttention overlaps cache transfer with computation \citep{gao2024cachedattention}, and LMCache supports cache movement between tiers \citep{liu2025lmcache}. These approaches make reuse more available. RadixFlow instead schedules the work left after the native state check.

\paragraph{Keeping complete hybrid state.}
Mamba, Mamba-2, and Gated Delta Networks require recurrent state at each reuse boundary \citep{gu2023mamba,dao2024ssm,yang2024gdn}. Marconi balances expected reuse and compute saving against hybrid-state memory cost \citep{pan2025marconi}; SGLang's unified cache tracks the components needed at a boundary \citep{sglang2026unified}. These policies concern state retention. RadixFlow uses the native state check \citep{vllm2026cache} to distinguish published state from a same-round dependency during selection.

\paragraph{Placing and ordering requests.}
Orca schedules at iteration level \citep{yu2022orca}, and Sarathi-Serve balances prefill and decode through chunking \citep{agrawal2024sarathi}. DistServe and Splitwise place the two phases on separate resources \citep{zhong2024distserve,patel2023splitwise}; FastServe and Llumnix adapt execution through preemption or migration \citep{wu2023fastserve,sun2024llumnix}. Prefix-aware Preble and $k$-LPM balance reuse against load and waiting \citep{srivatsa2025preble,dexter2025scheduling}. RadixFlow preserves the native request order and changes selection within one round when a same-round dependency is encountered.

\paragraph{Forming and computing batches.}
BatchLLM combines prefix sharing with token batching \citep{zheng2026batchllm}, while FastTree couples grouping with tree-aware kernels \citep{pan2025fasttree}. Dynamic batching collects work under a waiting bound \citep{triton2025batching}, which our timer controls isolate. FlashAttention, FlashAttention-2, FlashInfer, and vAttention improve the computation or layout of the resulting batch \citep{dao2022flashattention,dao2023flashattention2,ye2025flashinfer,prabhu2024vattention}. Because these decisions change per-round cost, we measure that cost rather than equating fewer rounds with proportional speedup.

\section{Additional Waiting and Capacity Analysis}
The full-attention Qwen3-8B (FA) isolates admission from the recurrent check. The check never fires, and Radix+Flex and RadixFlow have almost the same round count in every stream. Their aggregate reduction relative to Radix is about 27\%, with little change in mean latency. Its p50 instead rises from 3.24 to 3.63\,s under RadixFlow, so fewer rounds are not uniformly better for requests. Nemotron-9B exposes a different balance: admission removes 27\% of the rounds, and traversal removes another 8\% relative to Radix+Flex. At its lower offered load, Radix starts with only 2.91 rows per round out of 16; Flex raises this to 3.99 before traversal raises it to 4.38. This pattern is consistent with admission filling sparsely occupied rounds. Its host cost stays nearly constant, which allows the reduced round count to translate more directly into time saving. Appendix~\ref{app:savedtraces} gives the regime and distribution summaries.

Equation~\ref{eq:hostdecomposition} separates the credit for eliminating rounds from the cost shift in the remaining rounds. We use Radix as index 0 and RadixFlow as index 1. On Qwen3.5-9B, the cost shift absorbs about 60\% of the gross credit (Figure~\ref{fig:hostdecomposition}). On Nemotron-9B, per-round cost changes little, so more of the round reduction survives as a reduction in total host-wrapper time. A shape-adjusted analysis of tokens and rows follows the same pattern: batch shape explains much of the Qwen variation, with smaller residual policy associations (Appendix~\ref{app:savedtraces}). Direct attribution to state restoration or queue scanning requires the component measurements in Appendix~\ref{app:componentcontract}.

\paragraph{Waiting and tail latency.}
Client latency includes service-side waiting. Offered-arrival latency also includes sender lag when the client pool is full. For Qwen3.5-27B, RadixFlow lowers mean sender lag from 721 to 697\,s, in addition to lowering post-send latency from 164 to 159\,s. These are distinct waiting intervals, and their tail quantiles are measured on each request rather than added together. On Qwen3.5-9B, p95 falls from 73.3 to 70.6\,s and p99 from 75.2 to 72.6\,s; on Nemotron-9B, they fall from 15.8 to 13.3\,s and from 18.8 to 16.1\,s. The full-attention control has nearly unchanged tails; Appendix~\ref{app:savedtraces} covers all six policies.

\paragraph{How close are the rounds to the capacity bound?}
For each stream and order, we compare actual rounds with $\max(\lceil W/T\rceil,\lceil V/S\rceil)$, using that policy's observed scheduled tokens $W$ and row visits $V$. On Qwen3.5-27B, RadixFlow reaches 1.13--1.14 times this capacity-only bound across the three streams. On Qwen3.5-9B, the ratio is 1.44 in cold traffic but 4.18 in reuse traffic. High reuse leaves little token work, yet alignment, arrival times, and state dependencies still require rounds. The bound relaxes these constraints rather than describe an attainable online schedule.

\section{Worker Cost and Output Audit}
\label{app:outputaudit}
\begin{table}[t]\centering\small\begin{tabular}{lr}\toprule
Model & Decoded agreement (\%)\\\midrule
Qwen3.5-0.8B & 98.477 \\
Qwen3.5-4B & 99.850 \\
Qwen3.5-9B & 99.680 \\
Qwen3.5-27B & 99.995 \\
Nemotron-9B & 99.688 \\
Qwen3-8B (FA) & 99.441 \\
\bottomrule\end{tabular}\caption{Exact decoded first-token agreement between RadixFlow and Radix on 37,224 paired branches per model.}\label{tab:outputagreement}\end{table}

\begin{table*}[t]\centering\small
\begin{tabular}{lrrrrrr}\toprule
Model & Radix & Timer5 & Timer10 & Flex & RadixFlow & +Frag\\\midrule
Qwen3.5-0.8B & 52.95 & 52.99 & 53.09 & 53.64 & 54.24 & 53.94 \\
Qwen3.5-4B & 82.63 & 83.26 & 84.14 & 83.42 & 92.95 & 91.69 \\
Qwen3.5-9B & 112.85 & 116.78 & 116.67 & 113.38 & 145.50 & 144.16 \\
Qwen3.5-27B & 729.94 & 829.01 & 836.45 & 738.22 & 1262.95 & 1246.66 \\
Nemotron-9B & 53.06 & 53.82 & 53.10 & 53.69 & 53.87 & 54.31 \\
Qwen3-8B (FA) & 26.60 & 27.55 & 27.73 & 27.42 & 27.54 & 27.42 \\
\bottomrule\end{tabular}
\caption{Mean host-wrapper milliseconds per nonempty global worker round, retained for all six methods.}\end{table*}
The saved HTTP responses record one completed output token per branch. We compare decoded text exactly, with no normalization of whitespace or special-token text. Every policy has 37,224 paired branches per model. Agreement with Radix ranges from 98.477\% to 99.995\% for RadixFlow; the audit retains differing strings. The audit quantifies sensitivity to the scheduling configuration. Repeat-versus-repeat controls would separate native variability from policy effects. A downstream quality comparison uses candidate likelihoods, a fixed tie rule, and gold answers; the present text-agreement metric measures serving-output stability.

\section{Baseline Audit and Replay Sensitivity}
\label{app:baselineaudit}
\paragraph{Common overlay versus stock vLLM.}
Radix is the native-selection arm of the common experimental backend. All six arms execute registration/index hooks and tracing, and use the same copy-coverage guard. Their paired differences measure scheduling on a common backend. A stock vLLM+APC arm must remove the overlay entirely while retaining the same model, input tokens, arrival trace, and resource settings. The table reports the six common-backend policies named in its rows.

\paragraph{Alternative execution structures.}
Producer-first grouping executes a common producer before admitting its consumers, exchanging online packing freedom for an explicit group dependency. SGLang's prefix/fork programming model is another relevant comparison \citep{zheng2024sglang,sglang2026unified}. These alternatives require matched producer cost, cache capacity, and request-completion clocks, rather than comparing only consumer latency. These execution structures form a separate comparison axis from the six scheduling controls.

\paragraph{Timer eligibility and order.}
Timer5/10 register each branch before gating. A branch carries a deadline; peers are released when their earliest deadline expires or the row capacity is reached. The scheduler temporarily removes held branches and prepends them back after scanning executable work. In addition, the hold test can release peer timers while constructing the held list. Thus the control changes effective eligibility and can let a later branch execute before an earlier held one. The timer control therefore implements bounded admission with dynamic eligibility. An order-preserving timer control should snapshot eligibility before selection and record overtakes and deferral counts.

The 27B Timer10 tail split appears within every stream and both orders: p50 ranges from 10.53 to 11.80\,s and p99 from 910.83 to 1104.77\,s. It is therefore not created solely by pooling different streams. All requests completed. The code audit identifies eligibility changes consistent with the split; starvation attribution would require per-request deferral events. The reported Timer measurements use this eligibility rule.

\paragraph{The 4B mixed-stream result.}
The 61.2\% mean reduction persists in both assignment orders. Forward latency changes from 10.31 to 3.30\,s (68.0\%); reverse changes from 9.78 to 4.50\,s (54.0\%). The two orders are not independent timing repetitions. The benefit appears in both assignments, with its magnitude changing as the workload order changes.

\paragraph{Operating points and uncertainty.}
The 27B rate of 5 requests/s exceeds its observed drain-inclusive completion rate near 0.74 requests/s, producing sustained overload and sender backpressure. We use saturation here to describe that operating condition; the replay rate is not a load-sweep estimate of a throughput ceiling. Post-send and offered-to-completion latency are separate outcomes. Between-model comparisons also change row capacity and sometimes offered rate. The figure presents configurations, not a causal model-size trend.

Each frozen condition is run once. Forward/reverse order comparisons measure assignment sensitivity, not run-to-run variance. Host resources may be shared across services. Fixed-row, isolated-host repetitions at 50/80/95\% of independently measured native capacity would provide uncertainty for the small effects and establish a production operating envelope. CUDA-graph and asynchronous scheduling require separate matched runs; an asynchronous dependency must be identified by producer generation and completion fence, rather than a synchronous step number alone.

\end{document}